# Partially Specified Belief Functions

Serafín Moral, Luis M. de Campos
Departamento de Ciencias de la Computación e I.A.
Universidad de Granada
18071 - Granada - Spain
email: smc@robinson.ugr.es, lmcampos@ugr.es,

## Abstract

This paper presents a procedure to determine a complete belief function from the known values of belief for some of the subsets of the frame of discernment. The method is based on the principle of minimum commitment and a new principle called the focusing principle. This additional principle is based on the idea that belief is specified for the most relevant sets: the focal elements. The resulting procedure is compared with existing methods of building complete belief functions: the minimum specificity principle and the least commitment principle.

## 1  INTRODUCTION

If we have a variable, $X$, taking its values on a finite set, $U$, a belief function can be used to represent somebody's degree of support on the value of $X$ belonging to the different subsets of $U$. However, if we are working on a concrete case, we may find that the belief is not specified for all the subsets of $U$, but only for some of them. In general, the reasoning procedure needs a global belief defined for all the subsets. To build this global belief, the minimum specificity principle was proposed by Dubois and Prade, [Dubois-Prade 86a]. Intuitively speaking, this principle is similar to the maximum entropy principle for probability theory. Essentially, it says that when we have partial knowledge of a belief function, we have to select the least informative belief compatible with the available data.

The determination of the least specific belief function verifying a set of constraints may be done by solving a classical linear programming problem. However, the number of variables is large ($2^n$, where $n$ is the number of elements in $U$) and the resolution may involve a lot of calculations. To cope with this problem, it would be good to determine direct methods to calculate a belief function, or better still, its mass assignment, without resorting to optimization problems. In this paper we consider a particular case (when the family of subsets is closed under intersection) in which the solution is direct. We also propose a method to reduce the number of variables of the associated linear programming problem.

In general, to build a least specific belief function compatible with some known beliefs is not an easy task. Even a simpler question: Is there a belief function compatible with available data? is, in general, difficult to answer without doing linear programming. Lemmer and Kyburg, [Lemmer-Kyburg 91], study this problem when what we know are the belief-plausibility intervals for singletons. They propose an algorithm to calculate a belief function compatible with the available data, however, in general, it is not the least informative one.

Hsia, [Hsia 90], has considered the principle of minimum commitment as a basis for building complete belief functions from incomplete data. However, he does not consider how to effectively calculate the resulting belief function. This principle is a stronger version of the minimum specificity principle: when there is a least committed belief there is a least specific belief and the two are equal, but the opposite is not true.

In this paper, we introduce a new principle, which may help to solve some cases in which the minimum commitment criterium does not provide a solution. It will be called the focusing principle and its underlying idea is that when somebody is giving us his beliefs for some of the subsets of $U$, he is not choosing the subsets in an arbitrary way. Our hypothesis is that people specify their beliefs for the relevant sets.

When this principle can be applied it will produce an explicit expression for the belief function and the mass assignment. Its calculation is therefore much simpler than the resolution of the linear programming problem associated with the minimum specificity principle.

The focusing principle will not be applicable in all situations. We shall give the conditions under which this principle will produce a complete belief and propose alternative procedures for the cases in which they are not verified.

The basic concepts of belief functions and the preliminary results are presented in section 2. In section 3 we study the existing principles: minimum specificity and least commitment, and we give methods to reduce the number of variables associated to the minimum specificity principle. In section 4 we introduce the focusing principle and compare its behaviour with the results of applying the other principles. Finally section 5 is devoted to the conclusions.

## 2  BASIC CONCEPTS AND RESULTS

A belief function on a finite set, $U$, is a mapping, [Dempster 67, Shafer 76, Smets 88],

$$Bel : 2^U \to [0,1]$$

verifying the following properties

1. $Bel(\emptyset) = 0, \quad Bel(U) = 1$
2. $\forall n \geq 1, \forall A_1, \ldots, A_n \subseteq U$

$$Bel(A_1 \cup \ldots \cup A_n) \geq \sum_{\substack{I \subseteq \{1,\ldots,n\} \\ I \neq \emptyset}} (-1)^{|I|+1} Bel(\bigcap_{i \in I} A_i)$$

where $|I|$ is the number of elements in the set $I$.

$Bel(A)$ is the amount of support for the assertion: the true value belongs to the set $A$.

The function $Pl$ defined by

$$Pl(A) = 1 - Bel(\overline{A}) \qquad (1)$$

is called the plausibility associated with $Bel$. $Pl(A)$ is the amount of support that is not given to $\overline{A}$. $Bel$ is a kind of lower certainty value, whereas $Pl$ is an upper certainty value. In fact, we have $Bel(A) \leq Pl(A), \forall A \subseteq U$.

A mass assignment is a mapping

$$m : 2^U \to [0,1]$$

verifying

1. $m(\emptyset) = 0$
2. $\sum_{A \subseteq U} m(A) = 1$

It can be shown that a function, $Bel$, is a belief function if and only if there is a mass assignment, $m$, such that

$$Bel(A) = \sum_{B \subseteq A} m(B) \qquad (2)$$

The mass assignment can be calculated from the belief function on an explicit way. We have the following formula, [Shafer 76]:

$$m(A) = \sum_{B \subseteq A} (-1)^{|B-A|} Bel(B) \qquad (3)$$

In this way, given a belief function, we can talk of its associated mass assignment.

A subset, $A \subseteq U$, is said to be focal if and only if $m(A) > 0$. The set of focal elements of a mass assignment will be denoted by $\mathcal{F}$.

One of the advantages of using the mass assignment is that in most cases the number of focal subsets is small, and then calculations can be carried out in an efficient way. It is only necessary to know the mass assignment on the focal elements. In this section we shall prove a similar result for the belief values: We shall give a simple procedure to calculate the values of belief for the non focal elements from the values of belief of focal elements.

If $Bel_1$ and $Bel_2$ are two belief functions, $Bel_1$ is said to be less committed than $Bel_2$ if and only if, [Dubois-Prade 86b],

$$Bel_1(A) \leq Bel_2(A), \quad \forall A \subseteq U \qquad (4)$$

The idea underlying less commitment relationship (also called weak ordering, [Dubois-Prade 86b]) is that if $Bel_1$ is less committed than $Bel_2$ then the first is less informative than the second. In fact, if (4) is verified then we have

$$[Bel_2(A), Pl_2(A)] \subseteq [Bel_1(A), Pl_1(A)] \quad \forall A \subseteq U \quad (5)$$

That is, the belief-plausibility intervals associated to $Bel_2$ are more specific than the intervals associated to $Bel_1$.

If $Bel$ is a belief function and $m$ its mass assignment, the degree of specificity of $Bel$ is the value, [Yager 83]

$$S(Bel) = \sum_{A \subseteq U} \frac{m(A)}{|A|} \qquad (6)$$

There are other measures of specificity of a belief function, [Dubois-Prade 87, Campos et al. 90], but they have similar behaviour and there are no fundamental differences in the resulting principles. You may obtain different results using different specificity measures, but the intuitive basis of the associated principles is always the same.

In the following we give some technical definitions.

If $\mathcal{A} \subseteq 2^U$ is a family of subsets of $U$, then if $A, B \in \mathcal{A}$, $A \wedge_\mathcal{A} B$ is defined as the set of maximal elements of

$$\{C \in \mathcal{A} \mid C \subseteq A, C \subseteq B\}$$

with respect to the inclusion relation.

That is $D \in A \wedge_{\mathcal{A}} B$ if and only if

1. $D \in \mathcal{A}$
2. $D \subseteq A, D \subseteq B$
3. If $E \in \mathcal{A}$, $E \subseteq A, E \subseteq B$, and $D \subseteq E$, then $D = E$.

If $A_1, \ldots, A_n \in \mathcal{A}$, $\bigwedge_{i=1}^{n} A_i$ is defined in an analogous way. It is equal to the set of maximal elements of:

$$\{C \in \mathcal{A} \mid C \subseteq A_i, i = 1, \ldots, n\}$$

If $\mathcal{A} = 2^U$, then $A \wedge_{\mathcal{A}} B = \{A \cap B\}$. However, in general we get a family of elements of $\mathcal{A}$, which may be empty or have one or more elements.

The following propositions are the basis of our work. The proofs are technical and will be omitted.

**Proposition 1** *If Bel is a belief function and $m$ its associated basic probability assignment, then $\forall B_1, \ldots, B_k \subseteq U$,*

$$Bel(B_1 \cup \ldots \cup B_k) - \sum_{\substack{I \subseteq \{1,\ldots,k\} \\ I \neq \emptyset}} (-1)^{|I|+1} Bel(\bigcap_{i \in I} B_i)$$

$$= \sum_{A \in \mathcal{H}_k} m(A) \quad (7)$$

*where $\mathcal{H}_k = \{A \mid A \subseteq B_1 \cup \ldots \cup B_k, A \not\subseteq B_i, \forall i \in \{1, \ldots, k\}\}$.*

**Proposition 2** *If Bel is a belief function and $\mathcal{H}$ is a set of subsets of $U$ containing all the focal elements ($\mathcal{F} \subseteq \mathcal{H}$), then if $A \subseteq U$ and $\mathcal{T}_A = \{B_1, \ldots, B_k\}$ is the set of maximal elements of*

$$\{B \mid B \in \mathcal{H}, B \subset A, B \neq A\}$$

*with respect to the inclusion relation, then*

$$m(A) = Bel(A) - \sum_{\substack{I \subseteq \{1,\ldots,k\} \\ I \neq \emptyset}} (-1)^{|I|+1} Bel(\bigcap_{i \in I} B_i)$$

If we have a family $\mathcal{H}$ such that $\mathcal{F} \subseteq \mathcal{H}$, and $\{B_1, \ldots, B_k\} \subseteq \mathcal{H}$, we define the belief of $\{B_1, \ldots, B_k\}$ in a recursive way,

$$BEL_{\mathcal{H}}(\emptyset) = 0$$

$$BEL_{\mathcal{H}}(\{B_1\}) = Bel(B_1)$$

If $k \geq 2$,

$$BEL_{\mathcal{H}}(\{B_1, \ldots, B_k\}) =$$

$$\sum_{\substack{I \subseteq \{1,\ldots,k\} \\ I \neq \emptyset}} (-1)^{|I|+1} BEL_{\mathcal{H}}(\bigwedge_{i \in I} B_i)$$

where $\wedge$ is carried out with respect to the family $\mathcal{H}$.

The calculation of $BEL_{\mathcal{H}}(\{B_1, \ldots, B_k\})$ always stops because each time we perform $\bigwedge_{i \in I} B_i$ we obtain smaller sets from $\mathcal{H}$. This ensures that we shall arrive to the calculation of $BEL_{\mathcal{H}}$ for the empty set or for unitary sets. In this case, $BEL_{\mathcal{H}}$ is calculated directly from $Bel$. However the calculation may be hard. A more efficient calculation can be carried out through the following equality:

$$BEL_{\mathcal{H}}(\{B_1, \ldots, B_k\}) = \sum_{A \in \mathcal{B}_k} m(A) = \sum_{A \in \mathcal{B}'_k} m(A) \quad (8)$$

where

$$\mathcal{B}_k = \{A \in 2^U \mid \exists i \in \{1, \ldots, k\}, \text{ such that } A \subseteq B_i\}$$

$$\mathcal{B}'_k = \{A \in \mathcal{H} \mid \exists i \in \{1, \ldots, k\}, \text{ such that } A \subseteq B_i\}$$

This value of belief, $BEL_{\mathcal{H}}(\{B_1, \ldots, B_k\})$, only depends on the value of the belief function, $Bel$, on the subsets in the family $\mathcal{H}$.

**Example 1** *If $U = \{u_1, u_2, u_3, u_4\}$ and $\mathcal{H} = \{\{u_1, u_2, u_3\}, \{u_2, u_3, u_4\}, \{u_3\}\}$ then,*

$$BEL_{\mathcal{H}}(\{\{u_1, u_2, u_3\}, \{u_2, u_3, u_4\}\}) =$$

$$Bel(\{u_1, u_2, u_3\}) + Bel(\{u_2, u_3, u_4\}) - Bel(\{u_3\}) =$$

$$m(\{u_1, u_2, u_3\}) + m(\{u_2, u_3, u_4\}) + m(\{u_3\})$$

**Proposition 3** *If Bel is a belief function and $\mathcal{H}$ is a set of subsets of $U$ containing all the focal elements ($\mathcal{F} \subseteq \mathcal{H}$), then if $A \subseteq U$ and $\mathcal{T}_A = \{B_1, \ldots, B_k\}$ is the set of maximal elements of*

$$\{B \mid B \in \mathcal{H}, B \subset A, B \neq A\}$$

*with respect to the inclusion relation, then*

$$m(A) = Bel(A) - \sum_{\substack{I \subseteq \{1,\ldots,k\} \\ I \neq \emptyset}} (-1)^{|I|+1} BEL_{\mathcal{H}}(\bigwedge_{i \in I} B_i)$$

In last proposition, a more efficient calculation of the masses can be carried out, if we use expression (8) and follow and order in which the masses of the sets included in $A$ are calculated before the mass of set $A$.

The following proposition says that the belief of the non focal elements can be obtained from the belief of the focal ones.

**Proposition 4** *If Bel is a belief function and $\mathcal{F}$ is the set of its focal elements, then if $A \subseteq U, A \notin \mathcal{F}$ and $\mathcal{T}_A = \{B_1, \ldots, B_k\}$ is the set of maximal elements of*

$$\{B \mid B \in \mathcal{F}, B \subset A, B \neq A\}$$

*with respect to the inclusion relation, then*

$$Bel(A) = \sum_{\substack{I \subseteq \{1,\ldots,k\} \\ I \neq \emptyset}} (-1)^{|I|+1} BEL_\mathcal{F}(\bigwedge_{i \in I} B_i)$$

The calculation of $BEL_\mathcal{F}$ by using (8) and Proposition 3 to calculate the mass of the focal elements included in $A$ is more efficient.

## 3  THE MINIMUM SPECIFICITY AND LEAST COMMITMENT PRINCIPLES

Let us assume we have a finite set $U$ and that we know the following values of belief for some of the subsets of $U$,

$$Bel(A_i) = a_i, \qquad i = 1, \ldots, n \qquad (9)$$

The family of sets $\{A_1, \ldots, A_n\}$ for which we know the belief will be called $\mathcal{H}$. We shall consider that the complete frame $U$ and the empty set always belong to $\mathcal{H}$, because we know their beliefs: $Bel(\emptyset) = 0, Bel(U) = 1$.

We try to build a complete belief function, assigning a value of belief to every subset of $U$. To solve this problem we can apply the minimum specificity principle, [Dubois-Prade 86a]. According to it we have to consider the belief function with minimum specificity among those verifying conditions (9). Such a belief function can be calculated by solving the following linear programming problem:

$$\begin{aligned}
Min \quad & \sum_{A \subseteq U} \frac{m(A)}{|A|} \\
s.t. \quad & \sum_{A \subseteq A_i} m(A) = a_i, \quad i = 1, \ldots, n \\
& m(A) \geq 0
\end{aligned} \qquad (10)$$

The optimum of this problem provides the mass assignment, $m$, of the belief verifying the minimum specificity principle. It is important to notice that the solution is not always unique, and so some additional procedure should be applied to select a belief function. Our proposal is to apply what we call *the symmetry principle*. Intuitively, this principle says that if there are several possible solutions we should look for an intermediate solution among the extreme ones. As the set of solutions to (10) is a convex set, the center of gravity of this convex set, given by the arithmetic mean of its extreme points, looks appropriate to be proposed as the complete belief.

Another important point for problem (10) is that the number of variables of this problem is $2^n$, where $n$ is the number of elements of $U$. This may make its resolution unfeasible even for moderate values of $n$.

**Example 2** *Let $U = \{u_1, u_2, u_3\}$ be the frame of discernment, and let us suppose that we know the belief of set $\{u_1, u_2\}$ which is equal to 0.5. In this case there is one and only one solution to problem (10), which is the mass assignment given by,*

$$m(\{u_1, u_2\}) = 0.5, \quad m(U) = 0.5,$$
$$m(A) = 0 \quad otherwise$$

*If for the same frame we know the following belief values:*

$$Bel(\{u_1, u_2\}) = 0.6, \quad Bel(\{u_2, u_3\}) = 0.7$$

*then the minimum specificity principle gives rise to the following mass assignment:*

$$m(\{u_2\}) = 0.3, \quad m(\{u_1, u_2\}) = 0.3,$$
$$m(\{u_2, u_3\}) = 0.4$$

*If what we know are the following belief values:*

$$Bel(\{u_1, u_2\}) = 0.5, \quad Bel(\{u_2, u_3\}) = 0.5$$
$$Bel(\{u_1, u_3\}) = 0.5$$

*then, there is not a unique mass assignment with minimum specificity. The three following mass assignments and their convex combinations have minimum specificity:*

$$\begin{aligned}
m_1(\{u_2\}) &= 0.5, & m_1(\{u_1, u_3\}) &= 0.5 \\
m_2(\{u_1\}) &= 0.5, & m_2(\{u_2, u_3\}) &= 0.5 \\
m_3(\{u_3\}) &= 0.5, & m_3(\{u_1, u_2\}) &= 0.5
\end{aligned}$$

*Applying the symmetry principle, we get the following mass:*

$$m(\{u_1\}) = m(\{u_2\}) = m(\{u_3\}) = 0.5/3$$
$$m(\{u_1, u_2\}) = m(\{u_2, u_3\}) = m(\{u_1, u_3\}) = 0.5/3$$

Another principle that has been proposed to solve the problem of underspecification of a belief function is the least commitment principle. It was proposed by Hsia, [Hsia 90]. According to it, we have to select the least committed belief among those compatible with the given conditions. This is a kind of cautious minimum specificity principle. As is well known, [Dubois-Prade 86b, Yager 83], if $Bel_1$ is less committed than $Bel_2$, then the specificity of $Bel_1$ is smaller than the specificity of $Bel_2$, but the the inverse is not always true. So, if there is a least committed belief function, then there is a belief function with minimum specificity and both are the same. However, the inverse is not true.

We do not know any general algorithm to find the least committed belief function when one exists. This is one great disadvantage of this principle. However, it is possible to solve some particular cases.

**Example 3** *Let us review the cases of example 2 in the light of this principle.*

*In the first case we know the belief of the set $\{u_1, u_2\}$ which is equal to $0.5$. Applying this principle we obtain the same solution as above.*

*In the second case, when we know the belief values:*

$$Bel(\{u_1, u_2\}) = 0.6, \quad Bel(\{u_2, u_3\}) = 0.7$$

*once again, we obtain the same solution.*

*In the third case, we know the values,*

$$Bel(\{u_1, u_2\}) = 0.5, \quad Bel(\{u_2, u_3\}) = 0.5,$$
$$Bel(\{u_1, u_3\}) = 0.5$$

*There is no least committed belief function. We could also apply the symmetry principle, to select a belief function among those which are minimal with respect to the less commitment relationship. However, in this case, this principle presents an additional difficulty. The set of minimal beliefs is not convex, in general, and there is not a direct way to apply it.*

One disadvantage of the minimum specificity principle as opposed to the least commitment principle is that the first is sensitive to the change of granularity in the frame of discernment and the second is not, as the following example shows.

**Example 4** *Let us assume that in the last case of example 2, $u_3$ has been split in two cases: $v_1$ and $v_2$. Our new frame is $\{u_1, u_2, v_1, v_2\}$, and our knowledge of the beliefs is given by:*

$$Bel(\{u_1, u_2\}) = 0.5, \quad Bel(\{u_2, v_1, v_2\}) = 0.5,$$
$$Bel(\{u_1, v_1, v_2\}) = 0.5$$

*Now there is one and only one belief with minimum specificity, given by*

$$m(\{u_1, u_2\}) = 0.5, \quad m(\{v_1, v_2\}) = 0.5$$

*We observe that, by changing the granularity of the frame we change the solution. In relation with the least commitment principle, we have the same situation as above. There is not a least committed solution.*

In [Dubois-Prade 86a], Dubois and Prade show that if the sets for which we know the beliefs are $A_1$ and $A_2$, then the optimum to the problem (10) must have a set of focal elements included in the family $\{A_1, A_2, A_1 \cap A_2, U\}$. This is a very interesting result, because in this case the number of variables of the linear programming problem we have to solve is small, and the efficiency problems disappear. In the rest of this section we generalize this result to the case in which we know the beliefs for a general family $\mathcal{H}$ (including the frame $U$ and the empty set).

First, we consider the case in which this family $\mathcal{H}$ is closed for the intersection. In the following proposition we give a necessary and sufficient condition for the existence of the optimum, with an explicit expression for it. The proof is based on Proposition 2, and will not be given here.

**Proposition 5** *If we know the beliefs for a family of sets $\mathcal{H}$ which is closed under intersection, that is*

$$\forall A, B \in \mathcal{H}, A \cap B \in \mathcal{H}$$

*then there is a complete belief function in $U$ with these values of beliefs if and only if for every $A \in \mathcal{H}$, if $\mathcal{T}_A = \{B_1, \ldots, B_k\}$ is the set of maximal elements of*

$$\{B \mid B \in \mathcal{H}, B \subset A, B \neq A\}$$

*with respect to the inclusion relation, then*

$$Bel(A) \geq \sum_{\substack{I \subseteq \{1,\ldots,k\} \\ I \neq \emptyset}} (-1)^{|I|+1} Bel(\bigcap_{i \in I} B_i)$$

*Furthermore, in this case the optimum to the problem (10) is also the least committed belief and can be expressed in the following way:*

*If $A \subset U$, then*

- *If $A = A_i \in \mathcal{H}$, $Bel(A) = Bel(A_i) = a_i$ (the known belief)*

- *If $A \notin \mathcal{H}$, then if $\mathcal{T}_A = \{B_1, \ldots, B_k\}$ is as above*

  *If $m \geq 2$,*

  $$Bel(A) = \sum_{\substack{I \subseteq \{1,\ldots,k\} \\ I \neq \emptyset}} (-1)^{|I|+1} Bel(\bigcap_{i \in I} B_i)$$

In the above proposition $\mathcal{T}_A$ can never be empty because we assume that the empty set belongs to $\mathcal{H}$. Another interesting point is that it can be shown that the set of focal elements of the optimal belief $Bel$ is included in $\mathcal{H}$. A consequence of this observation is the following proposition.

**Proposition 6** *If we know the values of belief for a family $\mathcal{H}$ including the complete frame $U$ and the empty set, then if a feasible solution for the problem (10) exists, the optimum to such a problem has a family of focal elements, $\mathcal{F}$, included in $\widehat{\mathcal{H}}$, where*

$$\widehat{\mathcal{H}} = \{C \mid C = B_1 \cap \ldots \cap B_k, B_i \in \mathcal{H}\}$$

*is the clausure os $\mathcal{H}$ under intersection.*

If we know the values of belief for $\{A_1, A_2, A_3\}$, then we can can apply the above proposition to $\mathcal{H} = \{A_1, A_2, A_3, U, \emptyset\}$. It says that the focal elements of the least specific belief are in the family $\widehat{\mathcal{H}} = \{A_1, A_2, A_3, U, \emptyset, A_1 \cap A_2, A_1 \cap A_3, A_2 \cap A_3, A_2 \cap A_3 \cap A_3\}$. The empty set can be removed at this moment because it cannot be focal. This fact may simplify the solution to problem (10), because we do not have to consider the variables $m(A), A \notin \widehat{\mathcal{H}}$.

Finally the following proposition says that for the non existency of a belief it suffices that the conditions of proposition 5 are not verified for only one set $A \in \mathcal{H}$.

**Proposition 7** *If we know the beliefs for a family of sets $\mathcal{H}$, $A$ is an element of $\mathcal{H}$, $\mathcal{T}_A = \{B_1, \ldots, B_k\}$ is as in Proposition 5, verifying*

$$\forall I \subseteq \{1, \ldots, k\}, \bigcap_{i \in I} B_i \in \mathcal{H}$$

*and the following condition is verified*

$$Bel(A) < \sum_{\substack{I \subseteq \{1,\ldots,k\} \\ I \neq \emptyset}} (-1)^{|I|+1} Bel(\bigcap_{i \in I} B_i)$$

*then there is no complete belief function compatible with the given beliefs.*

## 4 THE FOCUSING PRINCIPLE

The focusing principle that we introduce in this paper to complete belief functions is based on the idea that when an expert expresses his beliefs for some of the subsets of $U$, he is not choosing the sets in an arbitrary way. We assume that he is expressing his beliefs for the relevant sets. Which are the relevant sets? According to Proposition 4, if a set is not focal, then its belief can be expressed as a function of the belief of the focal elements. In some way, the belief of the focal elements is the relevant knowledge we need to build a belief function. Then the focusing principle can be concreted in the following terms:

> When somebody is giving us his beliefs for a family of sets $\mathcal{H} = A_1, \ldots, A_n$, and we want to build a complete belief function, we should try to build a belief function in which the focal elements are in $\mathcal{H}$.

Our idea is that if we know the beliefs for a family $\mathcal{H}$ we should try to build the least committed belief function among those with a family of focal elements included in $\mathcal{H}$ and compatible with the given values.

The following proposition is a similar result to Proposition 5, but this time for the focusing principle. It says that if there is a belief function with focal elements included in $\mathcal{H}$, then there exists a least committed one, giving a direct expression for it.

**Proposition 8** *If we know the beliefs for a family of sets $\mathcal{H}$ then there is a complete belief function in $U$ with these values of beliefs and with a family of focal elements, $\mathcal{F}$ included in $\mathcal{H}$ if and only if for every $A \in \mathcal{H}$, if $\mathcal{T}_A = \{B_1, \ldots, B_k\}$ is as in proposition 5 then*

$$Bel(A) \geq \sum_{\substack{I \subseteq \{1,\ldots,k\} \\ I \neq \emptyset}} (-1)^{|I|+1} BEL_{\mathcal{H}}(\bigwedge_{i \in I} B_i)$$

*where $\wedge$ is carried out with respect to $\mathcal{H}$.*

*Furthermore, if the above conditions are verified then there is a least committed belief among those with focal elements included in $\mathcal{H}$ and compatible with the given values which can be expressed in the following way:*

*If $A \subset U$, then*

- *If $A = A_i \in \mathcal{H}$, $Bel(A) = Bel(A_i) = a_i$ (the known belief)*

- *If $A \notin \mathcal{H}$, then if $\mathcal{T}_A = \{B_1, \ldots, B_k\}$ is as above*

$$Bel(A) = \sum_{\substack{I \subseteq \{1,\ldots,k\} \\ I \neq \emptyset}} (-1)^{|I|+1} BEL_{\mathcal{H}}(\bigwedge_{i \in I} B_i)$$

In the above proposition, the values of $BEL_{\mathcal{H}}$ can be calculated from the values of $Bel$ in the family $\mathcal{H}$. So we give a direct procedure to caculate the complete least committed belief from the available data. This procedure is similar to the procedure used to calculate the least specific belief when the family $\mathcal{H}$ is closed under intersection. The advantage is that now, under the focusing principle, the family $\mathcal{H}$ is arbitrary.

When the conditions for the existence of a belief function with the focal elements included in $\mathcal{H}$ and the given values are verified, there is no problem. The above proposition provides a direct way of calculating the least committed belief (and therefore the least specific).

The problem is: What should we do if the existency conditions are not verified? The proposition says that there is no belief function with all the focal elements included in $\mathcal{H}$ and the given values. But there is nothing against the existence of a complete belief function with the focal elements included in $\widehat{\mathcal{H}}$. Our proposal will depend on the possibility of asking the expert for more values of belief. We consider two situations:

- *If the expert is available and we can ask for more values of belief.-* We ask for the belief of sets in $\widehat{\mathcal{H}}$ and not in $\mathcal{H}$. We do not try to obtain the belief values for all the subsets at the same time. Once we know the belief of one set, we try to verify the conditions of proposition 8. If they are verified we calculate the complete belief. If they are not verified we ask the belief for a new set from $\widehat{\mathcal{H}}-\mathcal{H}$. The procedure is the following:

  1. If the conditions of Proposition 8 are verified for $\mathcal{H}$ then
     - Calculate the complete belief
     - Exit
  2. While $\widehat{\mathcal{H}} - \mathcal{H} \neq \emptyset$ do 3-6
     3. Select $A \in \widehat{\mathcal{H}} - \mathcal{H}$
     4. $H = H \cup A$
     5. Apply Proposition 8
     6. If the conditions of Proposition 8 are verified then
        - Calculate the complete belief
        - Exit
  7. If the loop is finished without finding a complete belief, there is not a complete belief compatible with the available data.
  8. Exit

  The method of selection of the element $A \in \widehat{\mathcal{H}}-\mathcal{H}$ may be important. In order to find a complete belief as soon as possible, we should choose an element that might change the situation. In order to do that we propose the following rule: each time we start the loop is because the conditions of Proposition 8 are not verified. Assume that they are not verified for a set $B \in \mathcal{H}$ and that $\mathcal{T}_B = \{B_1, \ldots, B_k\}$. In this case it is useful to select a set $A = \bigcap_{i \in I} B_i$, where $I \subseteq \{1, \ldots, k\}$ that is not in $\mathcal{H}$. This set may make the conditions be verified for $B$.

  In order to keep the focal elements as close as possible to the original $\mathcal{H}$, the intersections $\bigcap_{i \in I_1} B_i$ with lower cardinal in $I$ should be introduced previously.

  If for a set $B$, all the sets $\bigcap_{i \in I} B_i$, where $I \subseteq \{1, \ldots, k\}$ are in $\mathcal{H}$ and the condition of Proposition 8 is not verified for it, then by Proposition 7 we are sure that there is no belief compatible with the available data, and we can stop the algorithm.

- *If the expert is not available.-* Then our proposal is to apply the minimum specificity criterium, but in a stepwise way. We know that the possible focal elements are in the set $\widehat{\mathcal{H}}$. However, if $\mathcal{H} = \{A_1, \ldots, A_n\}$, $\widehat{\mathcal{H}}$ can be expressed in the following way:

$$\widehat{\mathcal{H}} = \mathcal{H}^1 \cup \mathcal{H}^2 \cup \cdots \cup \mathcal{H}^n$$

where $\mathcal{H}^j = \{\bigcap_{i \in I} A_i \mid I \subseteq \{1, \ldots, n\}, |I| = j\}$

Then, instead of adding to $\mathcal{H}$ all the elements of $\mathcal{H}^2, \ldots, \mathcal{H}^n$ in one step, first we add the elements from $\mathcal{H}^2$, then the elements from $\mathcal{H}^3$, and so on. This is a weaker application of the focusing principle. We try to build a complete belief with focal elements as close as possible to the original ones. We are considering that the elements of $\mathcal{H}^2$ are closer than the elements from $\mathcal{H}^n$. The procedure is as follows:

1. Let $\mathcal{C} = \emptyset$
2. For $j = 1$ to $n$ do 3-5
   3. Let $\mathcal{C} = \mathcal{C} \cup \mathcal{H}^j$
   4. Solve the problem (10) with variables $m(A), A \in \mathcal{C}$
   5. If the problem has a solution
      - Give the optimum of the problem as the complete belief
      - Exit
6. If the loop ends without a solution there is no belief function compatible with the available data

**Example 5** *If we review the cases of Example 1, according to the focusing principle, we find that the focusing principle is directly applicable only in the first case producing the same solution as the minimum specificity principle. In the other two cases the weak application of the focusing principle (the expert is not available) produces the same solution as the minimum specificity criterium.*

**Example 6** *It seems that the focusing principle (applying the weaker version when it is not directly applicable) produces similar results to the minimum specificity principle. In a lot of situations this is true. However, there are some differences. Here we try to highlight the differences in their behaviour. Let us assume that $U = \{u_1, u_2, u_3, u_4, u_5\}$ and that we know the following beliefs:*

$$Bel(\{u_1, u_2\}) = 0.2, \ Bel(\{u_1, u_3\}) = 0.2, \\ Bel(\{u_1, u_4\}) = 0.2 \quad (11)$$

*In this case the minimum specificity principle produces a belief function whose mass assignment is:*

$$m_1(\{u_1\}) = 0.2, \quad m_1(U) = 0.8$$

*The focusing principle gives rise to the following belief function:*

$$m_2(\{u_1, u_2\}) = 0.2, \quad m_2(\{u_1, u_3\}) = 0.2,$$

$$m_2(\{u_1, u_4\}) = 0.2, \quad m_2(U) = 0.4$$

*This belief is more specific than the belief calculated with the minimum specificity principle, but our claim*

*is that it is closer to the way in which the beliefs were given. To specify the belief given by the mass, $m_1$, it would have been easier to specify simply that $Bel(\{u_1\}) = 0.2$. In that case the two principles would have produced the same result: $m_1$. Our idea is that to give $m_1$ we would not have expressed our beliefs as in (11). We would have used the simplest way: to say that $Bel(\{u_1\}) = 0.2$.*

## 5  CONCLUSIONS

In this paper we have studied the problem of building a complete belief function from the values of belief of some of the subsets of the frame. We have studied existing principles to carry out this task and introduced a new principle: the focusing principle.

This principle has the following important properties:

1. It applies the idea that belief is given for the relevant elements, then tries to build a belief function with focal elements among those for which we know the belief. This fact produces belief functions more in accordance with the expressed beliefs.

2. In general, the procedure is not complicated. When the principle is directly applicable, it produces an explicit expression for the complete belief. When it is not directly applicable and we consider the weaker version of the principle we have to solve several linear programming problems. However these problems are not independent, and we can start a problem, in which we have added a variable, in the point we left the former one. With this technique, in the worst case (we have to add all the variables corresponding to the sets in $\overline{\mathcal{H}} - \mathcal{H}$) we have a similar amount of computation to the application of the minimum specificity principle directly to the variables corresponding to $\widehat{\mathcal{H}}$.

We do not claim that the focusing principle should be applied in every possible situation. We think that this principle is more suitable for the cases in which the expert chooses the sets for which he is giving the beliefs. If on the contrary, we ask the expert for his beliefs on some sets determined by us, then the minimum specificity principle looks more appropriate. In that case, it would be interesting to determine optimal strategies to query the expert in such a way that the solution to the linear programming problem could be calculated in an efficient way.

If we have a belief function and we consider the values of the belief for the focal elements, then if we apply the focusing principle to them we obtain the same belief function. As a consequence, to represent a belief function we only have to know its values on the focal elements. The other values of belief can be calculated from these values. This open the possibility of using this representation to improve the efficiency of the calculations with belief functions.

## Acknowledgements

This work has been supported by the DGICYT under proyect PS59/152. We are very grateful to the anonymous referees for their valuable and useful comments and suggestions.